\documentclass[letterpaper, 10 pt, conference]{ieeeconf}  

\IEEEoverridecommandlockouts                              

\usepackage{amsfonts}       
\usepackage{nicefrac}       
\usepackage[dvipsnames]{xcolor}
\usepackage{microtype}      
\usepackage{graphicx}
\usepackage{caption}
\usepackage{subcaption}
\usepackage{booktabs}
\usepackage{algpseudocode}
\usepackage{xspace}
\usepackage{wrapfig}
\usepackage{ctable}
\usepackage{multirow}
\usepackage{varwidth}
\usepackage{adjustbox}
\usepackage{arydshln}
\definecolor{mydarkblue}{rgb}{0,0.08,0.45}
\usepackage[pagebackref=true,breaklinks=true,colorlinks,bookmarks=false,citecolor=mydarkblue]{hyperref}

\usepackage[capitalize]{cleveref}
\crefname{section}{Sec.}{Secs.}
\Crefname{section}{Section}{Sections}
\Crefname{table}{Table}{Tables}
\crefname{table}{Tab.}{Tabs.}

\newcommand{\ours}{SMERF}
\newcommand{\ourslong}{SD Map Encoder Representations from transFormers}
\newcommand\mypara[1]{\vspace{1.5mm}\noindent\textbf{#1}}

\newcommand{\dashrule}[1][black]{%
  \color{#1}\rule[\dimexpr.5ex-.2pt]{4pt}{.4pt}\xleaders\hbox{\rule{4pt}{0pt}\rule[\dimexpr.5ex-.2pt]{4pt}{.4pt}}\hfill\kern0pt%
}

\makeatletter
\DeclareRobustCommand\onedot{\futurelet\@let@token\@onedot}
\def\@onedot{\ifx\@let@token.\else.\null\fi\xspace}

\def\eg{\emph{e.g}\onedot} 
\def\ie{\emph{i.e}\onedot} 
 
\def\etc{\emph{etc}\onedot} \def\vs{\emph{vs}\onedot}

\makeatother

\title{\LARGE \bf Augmenting Lane Perception and Topology Understanding with Standard Definition Navigation Maps}

\author{Katie Z Luo$^{\dag}$, Xinshuo Weng$^\ddag$, Yan Wang$^\ddag$, Shuang Wu$^\ddag$, Jie Li$^\ddag$, \\Kilian Q Weinberger$^\dag$, Yue Wang$^{\S\ddag}$,
Marco Pavone$^{\P\ddag}$%
\thanks{$\dag$ Computer Information Sciences Department, Cornell University 
        {\tt\small \{kzl6, kqw4\}@cornell.edu}}%
\thanks{$\ddag$ NVIDIA
        {\tt\small \{xweng, yanwan, shwu, jieli\}@ nvidia.com}}%
\thanks{$\S$ Department of Computer Science, University of Southern California 
        {\tt\small yue.w@usc.edu}}%
\thanks{$\P$ Department of Aeronautics and Astronautics, Stanford University
        {\tt\small pavone@stanford.edu}}%
}

\begin{document}

\maketitle
\thispagestyle{empty}
\pagestyle{empty}

\begin{abstract}

Autonomous driving has traditionally relied heavily on costly and labor-intensive High Definition (HD) maps, hindering scalability. In contrast, Standard Definition (SD) maps are more affordable and have worldwide coverage, offering a scalable alternative. In this work, we systematically explore the effect of SD maps for real-time lane-topology understanding. We propose a novel framework to integrate SD maps into online map prediction and propose a Transformer-based encoder, \ourslong, to leverage priors in SD maps for the lane-topology prediction task. This enhancement consistently and significantly boosts (by up to 60\%) lane detection and topology prediction on current state-of-the-art online map prediction methods without bells and whistles and can be immediately incorporated into any Transformer-based lane-topology method. 
Code is available at \url{https://github.com/NVlabs/SMERF}.

\end{abstract}

\section{Introduction}
\label{sec:intro}

In order for autonomous driving and driver assistance systems to operate reliably, they need to be aware of the environment and scene elements in tremendous detail. Among others, accurate lane geometry, relational reasoning of lane graphs, and associating lanes to traffic lights and signs are paramount for vehicles to drive correctly. These tasks, known together as lane-topology \cite{wang2023openlanev2}, remain as challenges to deploying autonomous vehicles into the real world.

One solution to this challenge, where autonomous driving systems have found success, lies in deploying strictly within geo-fenced areas, where regions are annotated in detail in the form of High Definition (HD) maps \cite{li2022hdmapnet, sadat2020perceive,shi2023motion}.
HD maps typically include centimeter level map elements~\cite{liu_wang_zhang_2020} such as road boundaries, lane dividers, road markings, and traffic signs, as well as lane graphs and association of lanes to traffic signs. 
This precision mapping removes ambiguity from self-driving, making HD maps critical enablers for essentially all commercial robo-taxi services (e.g. Waymo, Cruise).  
In addition, HD maps also annotate areas like construction zones and pedestrian crossings to be high alert areas. 

While HD maps provide a solution for reliable self-driving, such maps are prohibitively expensive to obtain as each area needs to be painstakingly annotated by humans and continuously updated to reflect any changes in road conditions or ongoing daily constructions.
For these reasons, over-reliance on HD maps not only inhibits the scalability of self-driving, but also requires a large number of annotators at-the-ready to keep the maps constantly updated.

\begin{figure}
    \centering
    \includegraphics[width=\linewidth]{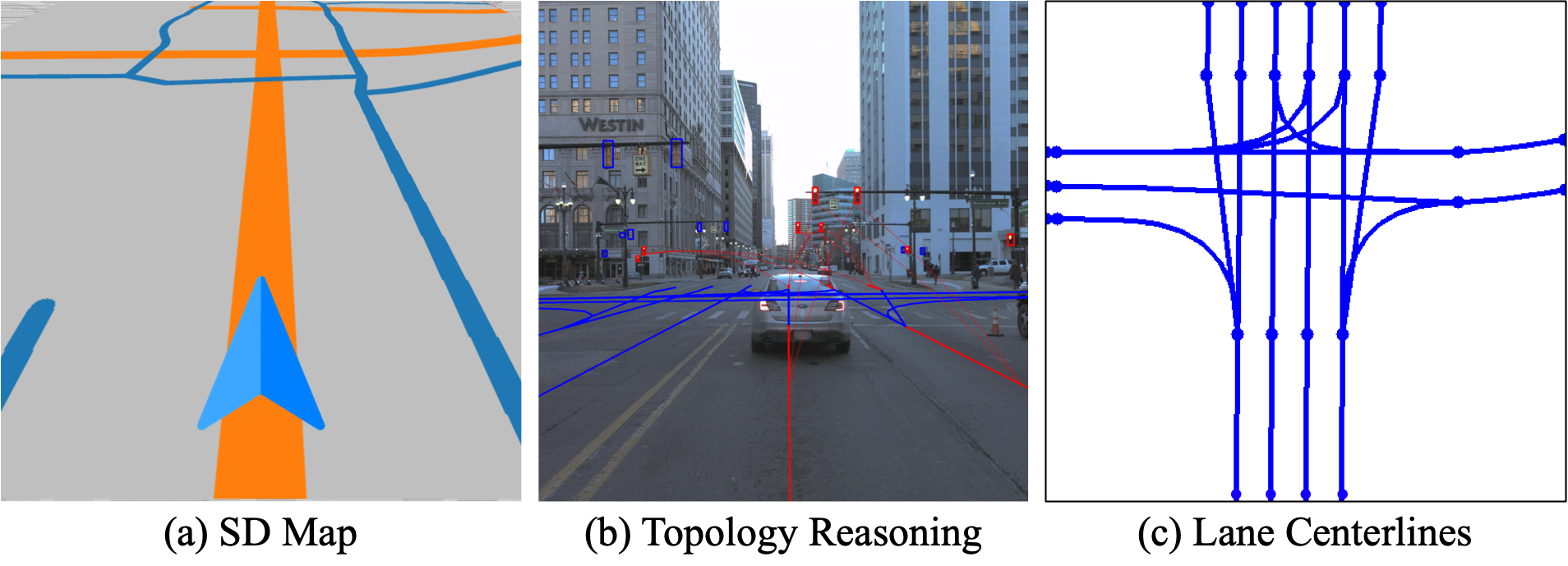}
    \caption{\textbf{Lane-Topology Reasoning.} Leveraging standard definition (SD) map (a) with prior information of the road-level topology, our work aims to improve lane centerline detection (c), lane-topology reasoning between lane centerlines, and traffic elements (b). In the SD map, {\color{orange}orange} lines and {\color{NavyBlue}teal} lines correspond to roads and pedestrian ways, respectively. 
    }
    \label{fig:teaser}
    \vspace{-\baselineskip}
\end{figure}

In contrast, Standard Definition (SD) maps are cheaper to obtain (\emph{e.g.,} crowdsourcing, aerial images) and are already available for much of the world's areas. 
Concretely, SD maps mark out road-level topology with metadata\footnote{Metadata can vary across different SD map providers, but it typically includes the type of the roads, the number of lanes of a road, the existence of a traffic sign, and the direction of the road.}, as opposed to the full semantic and geometric lane-level details in HD maps.
Such SD maps do not need to be updated as frequently as HD maps unless the road topology changes significantly (\eg new roads are constructed, or old roads are removed).

While being much cheaper and having broader coverage, SD maps include crucial information for road topology which can complement onboard cameras for lane-topology reasoning. In particular, when merging or exit roads are not visible in the camera images due to occlusion, an SD map can provide priors for more accurate downstream planning. Additionally, an SD map may provide priors over the existence of intersections before the self-driving car approaches. Such prior knowledge is helpful for long-horizon behavior planning, \emph{e.g.}, switching to a left lane early before making a left turn at the intersection. 

In this work, we explore the use of SD maps to improve online lane-topology reasoning in the absense of HD maps. 
We propose a novel and compellingly simple way of encoding the SD map in a Transformer encoder architecture \cite{vaswani2023attention} to learn feature representations that can be consumed in downstream lane-topology tasks. We name our method \emph{\ours{} (\ourslong{})}.

The framework for augmenting with SD maps is immediately applicable to any Transformer-based lane-topology methods, which currently dominate the state-of-the-art performance metrics \cite{li2023toponet,wu20231st,lu2023separated}. We demonstrate that adding SD maps as an additional source of information gives a boost in performance for lane-topology reasoning --- across \emph{all} available architectures. 
When used with the current best open-sourced lane-topology model \cite{li2023toponet}, lane detection and lane-topology prediction achieve state-of-the-art performance \textit{without any additional tuning}. This showcases the strong generalizability of \ours{} map representations, and is a testament to the inherent information present in SD maps for topology understanding. Our contributions are summarized as follows:
\begin{itemize}
    \item To our knowledge, we are the first work to systematically explore the utility of SD maps for lane-topology understanding. 
    \item We propose \ours{}, an SD map representation and Transformer based encoder model for lane-topology prediction. 
    \item We empirically demonstrate that our proposed method of incorporating SD maps significantly boosts the performance of \emph{all} lane-topology methods evaluated. 
\end{itemize}
\section{Related Work}
\label{sec:related}

\mypara{Online HD Map Prediction} has emerged to reduce the vast amount of human efforts in annotating and maintaining HD maps by predicting the HD maps on-the-fly while the car is in use. 
HDMapNet~\cite{li2022hdmapnet} pioneers learning-based models to predict map elements from onboard sensors, followed by a post-processing step to convert dense rasterized segmentations to vectorized map representations. To eliminate the need for hand-crafted post-processing, VectorMapNet~\cite{liu2022vectormapnet} and InstaGraM~\cite{shin2023instagram} introduced end-to-end models for vectorized HD map learning.
As the vectorized map learning typically involves key-point sampling along the polylines which can cause information loss, prior work~\cite{ding2023pivotnet, zhang2023online} proposed a pivot-based representation and differentiable rasterizer, tailored to corners and fine-grained geometry. 
Besides output representations, MapTR~\cite{liao2022maptr} and InsightMapper\cite{xu2023insightmapper} proposed a hierarchical query design and models map elements as a point set with a group of equivalent permutations. Recent work such as LaneGAP~\cite{liao2023lane} and TopoNet~\cite{li2023topology} also explored end-to-end approaches to learn the lane graph and model map element relationships.
In contrast to prior work which only uses onboard sensors as inputs, our work is the first to investigate the effect of SD maps to vectorized map learning and can be seamlessly employed to improve prior work.

\mypara{Learn to Fuse Prior for HD Map Learning.} Orthogonal to the above work, recent efforts capitalize on the fusion of prior information to improve the robustness and performance of online HD mapping. NMP~\cite{xiong2023neural} learns a neural representation and builds a global map prior from past traversals to improve online map prediction, while \cite{can2023prior} optimizes in the latent space to learn a global consistent prior of maps. \cite{gao2023complementing} supplements the onboard camera images with satellite images to deal with the long-range perception of HD maps. NeMO~\cite{zhu2023nemo} and StreamMapNet~\cite{yuan2023streammapnet} improve performance using temporal information by fusing frames from history, while Bi-Mapper~\cite{li2023bi} designed a multi-view fusion module to leverage priors in the perspective view. Our work is closely related to these methods, however, we leverage a different prior -- SD maps, that is much more compact in terms of storage and representation, and also widely accessible. 

\mypara{Transformers for 3D Perception.} Transformers have demonstrated their dominant performance in 3D perception from visual inputs. DETR3D~\cite{wang2022detr3d} leverages a Transformer architecture to connect 2D observations and sparse 3D predictions, enabling non-maximum suppression (NMS) free object detection. BEVFormer~\cite{li2022bevformer} extends DETR3D by transforming features from perspective views into a dense birds-eye view (BEV). PETR3D~\cite{liu2022petr} further incorporates 3D positional information into feature extraction, producing 3D position-aware features. In addition to camera images, recent methods leverage multi-modal inputs to complement a single modality. FUTR3D~\cite{chen2023futr3d} capitalizes on 3D-to-2D queries proposed by DETR3D to fuse features from multiple modalities. Our method is also a Transformer-based architecture that performs online mapping in an end-to-end fashion.

\section{Approach}
\label{sec:method}

\begin{figure*}[ht!]
    \centering
    \includegraphics[width=0.8\textwidth]{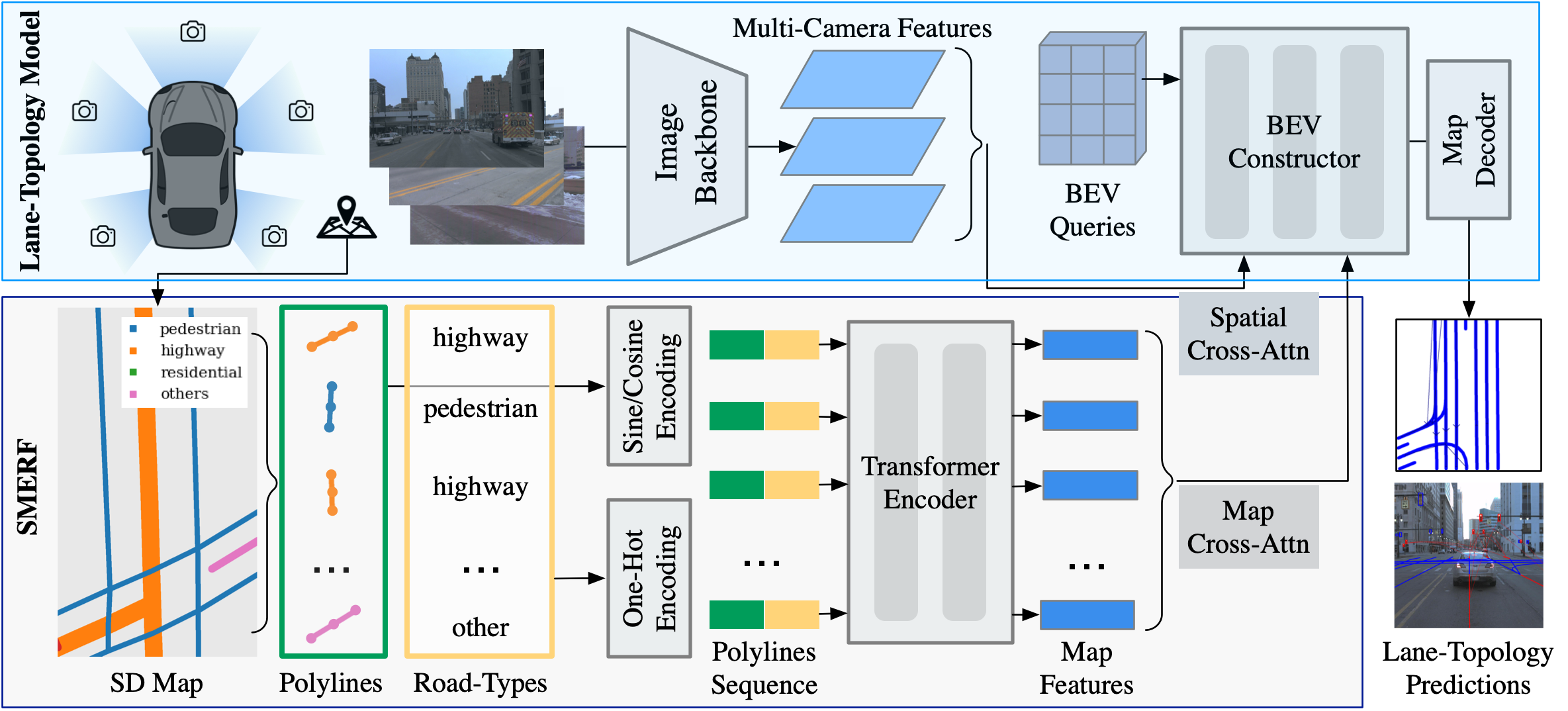}
    \caption{\textbf{Overall approach of \ours.} The SD map is queried at the ego vehicle's location as a prior for lane-topology reasoning. SD map is first converted into a polyline-sequence representation, then encoded by a Transformer encoder. Our method is amenable to any transformer-based lane-topology models via cross-attention with the SD map features.}
    \label{fig:model}
\end{figure*}

\mypara{Problem setup.} Following prior work \cite{wang2023openlanev2, li2023toponet}, we assume a multi-camera setup: the ego-vehicle is equipped with $C$ synchronized, multi-view cameras and their corresponding camera intrinsic and extrinsic parameters.
In addition, we have access to the SD maps and the ego-vehicle's 2D position and heading as a 3-DoF rigid transformation $G_p$ from a global positioning system (GPS) that is used to align the SD maps with onboard sensor inputs.
From these inputs, the task is to detect the \emph{lane centerlines} of the road and the \emph{traffic elements} of the scene such as the traffic lights and stop signs. Further, we infer the connectivity of the lane centerlines and how they relate to each traffic element. All pairwise relationships are represented as affinity matrices. 

The pipeline of \ours{} is shown in \autoref{fig:model}. The proposed \ours{} (lower half) augments an existing lane-topology model (upper half) with priors from SD maps in order to better detect lane centerlines and relational reasoning. 
Specifically, we first retrieve the SD map, which is encoded into a feature representation using a Transformer encoder. 
Then, we apply cross-attention between the SD map feature representation with the features from onboard camera inputs to construct the BEV features for lane detection and relational reasoning. The pipeline is trained end-to-end with the lane-topology model without requiring any additional training signals.

\subsection{SD Map Input}
\begin{figure}
    \centering
    \includegraphics[trim={0.3cm 0 0.2cm 0},clip,width=\linewidth]{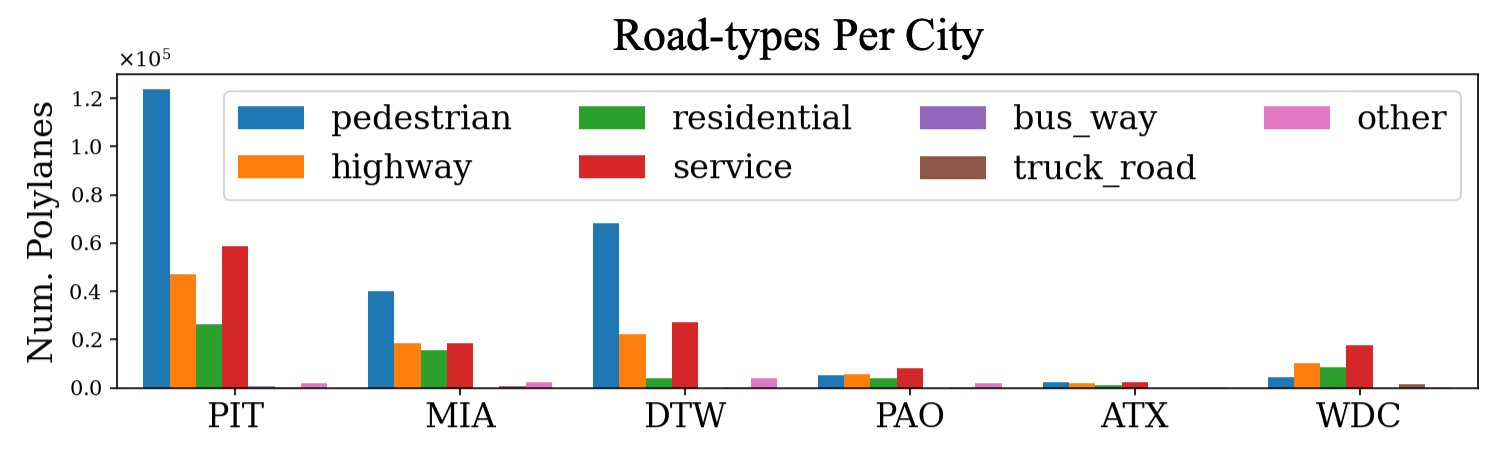}
    \caption{\textbf{Distribution of road types.} We visualize the road types obtained from OpenStreetMap corresponding to the frames in OpenLane-V2 dataset validation split. The colors distinguish by the cities the dataset is collected in.
    }
    \label{fig:analy-lanetype}
    \vspace{-\baselineskip}
\end{figure}

We obtain SD maps from OpenStreetMap (OSM)~\cite{bennett2010openstreetmap}, a crowd-sourced platform offering SD maps and geographical details of worldwide locations. Concretely, SD maps from OSM contain road-level topology (\ie road geometry and connectivity) and annotated type-of-road information for each road segment (\eg highway, residential roads, and pedestrian crossings). For every frame, we extract a local SD map from OSM based on the ego vehicle's position from $G_p$. The resulting SD map encompasses \( M \) polylines, where each polyline corresponds to a road segment. Notably, we transform the point location of the polylines to the ego vehicle's coordinates using $G_p$. Moreover, each polyline is further annotated with specific road-type labels. An in-depth analysis of the availability of the road types, along with their distribution patterns, is presented in \autoref{fig:analy-lanetype}. 

\subsection{Encoding Representation from SD Maps}
In order to encode the SD map in a form that can be consumed by a downstream lane-topology model, we introduce a polyline-sequence representation and a Transformer encoder to obtain our final map representation for the scene.

\mypara{Polyline Sequence Representation.}
Given the SD map of the scene, we evenly sample along each of the $M$ polylines for a fixed number of \( N \) points, denoted by \(\{(x_i, y_i)\}_{i=1}^N\).
We employ sinusoidal embeddings with varied frequencies to encode the polyline point locations. Sinusoidal embeddings enhance the sensitivity to positional variations. This sensitivity benefits the model, enabling it to effectively reason about the  structure of polylines. Consider a vertical polyline with a small curvature, characterized by closely similar \(y\)-axis values for all points. Directly inputting these
point coordinates into the model may result in an inadequate distinction of this curvature. However, with sinusoidal embeddings \cite{vaswani2023attention}, the distinction becomes pronounced, thus improving the model's interpretability of such features. Given a coordinate position \(p \in \{x_i, y_i\}\)
and an embedding dimension \(j \in \{1 \cdots d / 2\}\), the sinusoidal embedding can be formulated as:

\[ E(p, 2j) = \sin\left(\frac{p}{{T^{2j/d}}}\right), \]
\[ E(p, 2j+1) = \cos\left(\frac{p}{{T^{2j/d}}}\right), \]
where \(d\) is the dimension of the embedding, and \(T = 1000\) is the temperature scale. This enables the transformation of \((x_i,y_i)\) coordinates into their corresponding sinusoidal embeddings of dimension $d$.
In practice, we normalize each polyline's coordinates with respect to the BEV range into the range of \((0, 2\pi)\) prior to embedding them.

We use a one-hot vector representation for the road-type label with dimension $K$ for the main types of lanes present in OSM. This not only ensures that input values are normalized between 0 and 1, but also addresses cases where a road segment may fall into multiple road types.
Finally, we concatenate the polyline positional embeddings with the road type as one-hot vectors for the final polyline sequence representation with shape \( M \times (N \cdot d + K) \).

\mypara{Transformer Encoder of Map Features.}
Given the polyline sequence representation of the SD map, we wish to use a Transformer encoder \cite{vaswani2023attention} to learn a feature representation for the downstream lane-topology task.
We embed the polyline sequence with a linear layer, typical of Transformer encoder architectures. 
This ensures that the discrete, one-hot representation of the road-types can be meaningfully transformed into continuous space.
We then utilize \(L\) layers of multi-head self-attention to extract and encode the global geometric and semantic information from the SD map input. 
The resultant output has a shape of \( M \times H \), where \(H\) denotes the feature dimension produced by the self-attention layer.

\subsection{Lane-Topology Prediction with \ours}
The SD map representation from \ours{} can now be used by any Transformer-based lane-topology model.
With the release of the lane-topology task alongside the OpenLane-V2 dataset \cite{wang2023openlanev2}, the predominant paradigm for lane detection and relational reasoning models emerged consisting of an BEV Transformer encoder and a DeTR-based map decoder \cite{li2023toponet, wu20231st,lu2023separated}.
In this work, we propose to augment these state-of-the-art lane-topology models with features representations from the SD map. Current methods only leverage multi-view camera inputs, and have difficulty predicting in areas that are occluded or are far away. This is complimented by the additional SD map information, which allows the model to reason about these blind spots.

To make use of current state-of-the-art lane-topology models, \ours{} fuses the SD map features with the intermediate BEV feature representations by leveraging multi-head cross-attention. 
This method is compatible with nearly all transformer-based lane-topology models by applying cross-attention between the BEV feature queries and the SD map features in each intermediate layer of the model's encoder (\autoref{fig:model}, ``Map Cross-Attn"). 
In practice, we cross-attend the SD map features after each spatial cross-attention operation.
Thus, the fused BEV features include not only the 3D information derived from the images, but also the road-level geometric information extracted from the SD map. Subsequently, the lane-topology model decoder takes the SD map-augmented features as inputs to predict the lane centerlines, traffic elements, and affinity matrices for the association of lane centerlines and traffic elements.

\section{Experiments}
\label{sec:experiments}

\begin{table}[t]
\caption{\textbf{Performance on the OpenLane-V2 Dataset.} We report the performance of adding \ours{} to state-of-the-art lane-topology models, and we observe that \ours{} can significantly improve both the Baseline model and TopoNet.}
\label{tab:olv-main}
\begin{tabular}{@{}llccccc@{}}
\toprule
 &  & \multicolumn{1}{c}{DET\_l} & \multicolumn{1}{c}{TOP\_ll} & \multicolumn{1}{c}{DET\_t} & \multicolumn{1}{c}{TOP\_lt} & \multicolumn{1}{c}{OLS} \\ \midrule
Baseline  &  & 17.0 & 2.3 & 48.5 & 16.2 & 30.2 \\
Baseline+\ours{} &  & 26.8 & 3.9 & 48.9 & 19.2 & 34.8 \\ [-\jot]
  \multicolumn{7}{@{}c@{}}{\makebox[\linewidth]{\dashrule[black!40]}} \\[-\jot]
$\triangle$ Improvement &  & {\color[HTML]{34A853} 57.6\%} & {\color[HTML]{34A853} 73.1\%} & {\color[HTML]{34A853} 0.8\%} & {\color[HTML]{34A853} 18.6\%} & {\color[HTML]{34A853} 15.3\%} \\ \midrule
TopoNet &  & 28.2 & 4.1 & 44.5 & 20.6 & 34.5 \\
TopoNet+\ours{} &  & 33.4 & 7.5 & 48.6 & 23.4 & 39.4 \\ [-\jot]\multicolumn{7}{@{}c@{}}{\makebox[\linewidth]{\dashrule[black!40]}} \\[-\jot]
$\triangle$ Improvement &  & {\color[HTML]{34A853} 18.7\%} & {\color[HTML]{34A853} 83.8\%} & {\color[HTML]{34A853} 9.3\%} & {\color[HTML]{34A853} 13.6\%} & {\color[HTML]{34A853} 14.2\%} \\ \bottomrule
\end{tabular}
\end{table}

{\setlength{\tabcolsep}{2.6px}
\centering
\begin{table*}[t]

\centering
\caption{\textbf{Performance Breakdown.} We report the performance of baseline models as compared to \ours{} method, broken down by lane centerlines that are close \vs far, and non-intersection \vs intersections. Notice how improvements are greater at far away lanes and at intersections. We leave out DET\_t metrics for brevity, since we do not filter by traffic elements.}
\label{tab:olv-metrics-split}
\begin{tabular}{@{}llcccclcccclccccccccc@{}}
\toprule
 &  & \multicolumn{4}{c}{Close (0 - 25 m)} &  & \multicolumn{4}{c}{Far (25 - 50 m)} &  & \multicolumn{4}{c}{Non-Intersection/Connector} & \multicolumn{1}{l}{} & \multicolumn{4}{c}{Intersection/Connector} \\ \cmidrule(lr){3-6} \cmidrule(lr){8-11} \cmidrule(lr){13-16} \cmidrule(l){18-21} 
 &  & DET\_l & TOP\_ll & TOP\_lt & OLS &  & DET\_l & TOP\_ll & TOP\_lt & OLS &  & DET\_l & TOP\_ll & TOP\_lt & OLS & \multicolumn{1}{l}{} & DET\_l & TOP\_ll & TOP\_lt & OLS \\ \midrule
Baseline &  & 20.4 & 4.8 & 16.3 & 32.8 &  & 16.3 & 2.3 & 14.1 & 29.4 &  & 20.3 & 3.0 & 16.2 & 31.6 &  & 14.7 & 2.0 & 11.9 & 27.9 \\
Baseline+\ours{} &  & 22.4 & 6.6 & 18.4 & 35.0 &  & 26.3 & 3.9 & 17.5 & 34.2 &  & 29.2 & 5.3 & 19.3 & 36.3 &  & 21.2 & 3.7 & 14.9 & 32.0 \\ [-\jot]
  \multicolumn{21}{@{}c@{}}{\makebox[\linewidth]{\dashrule[black!40]}} \\[-\jot]
$\triangle$ Improvement &  & {\color[HTML]{34A853} 9.8\%} & {\color[HTML]{34A853} 38.1\%} & {\color[HTML]{34A853} 13.0\%} & {\color[HTML]{34A853} 6.8\%} & \multicolumn{1}{c}{{\color[HTML]{34A853} }} & {\color[HTML]{34A853} 61.0\%} & {\color[HTML]{34A853} 72.6\%} & {\color[HTML]{34A853} 23.9\%} & {\color[HTML]{34A853} 16.6\%} &  & {\color[HTML]{34A853} 43.4\%} & {\color[HTML]{34A853} 74.2\%} & {\color[HTML]{34A853} 19.0\%} & {\color[HTML]{34A853} 14.7\%} & {\color[HTML]{34A853} } & {\color[HTML]{34A853} 44.3\%} & {\color[HTML]{34A853} 87.8\%} & {\color[HTML]{34A853} 24.9\%} & {\color[HTML]{34A853} 14.6\%} \\ \midrule
TopoNet &  & 28.0 & 9.1 & 22.6 & 37.6 & \multicolumn{1}{c}{} & 26.5 & 4.4 & 18.2 & 33.6 &  & 32.7 & 8.1 & 21.2 & 37.9 &  & 23.7 & 4.6 & 16.0 & 32.4 \\
TopoNet+\ours{} &  & 32.4 & 12.6 & 24.3 & 41.5 & \multicolumn{1}{c}{} & 33.0 & 7.7 & 21.7 & 39.0 &  & 37.0 & 11.9 & 23.7 & 42.2 &  & 28.5 & 8.1 & 19.5 & 37.4 \\ [-\jot]
  \multicolumn{21}{@{}c@{}}{\makebox[\linewidth]{\dashrule[black!40]}} \\[-\jot]
$\triangle$ Improvement &  & {\color[HTML]{34A853} 15.7\%} & {\color[HTML]{34A853} 38.2\%} & {\color[HTML]{34A853} 7.6\%} & {\color[HTML]{34A853} 10.4\%} & \multicolumn{1}{c}{{\color[HTML]{34A853} }} & {\color[HTML]{34A853} 24.9\%} & {\color[HTML]{34A853} 76.8\%} & {\color[HTML]{34A853} 19.0\%} & {\color[HTML]{34A853} 16.0\%} &  & {\color[HTML]{34A853} 13.3\%} & {\color[HTML]{34A853} 46.6\%} & {\color[HTML]{34A853} 11.7\%} & {\color[HTML]{34A853} 11.3\%} & {\color[HTML]{34A853} } & {\color[HTML]{34A853} 20.3\%} & {\color[HTML]{34A853} 74.4\%} & {\color[HTML]{34A853} 22.1\%} & {\color[HTML]{34A853} 15.5\%} \\ \bottomrule
\end{tabular}
\end{table*}}

\mypara{Dataset and Evaluation Metrics.} We validate our approach on the OpenLane-V2 dataset \cite{wang2023openlanev2}, a large, real-world perception dataset for scene structure in autonomous driving. To the best of our knowledge, this is the only dataset providing ground truth to test lane and traffic detection, as well as topology relationships among lane centerlines and between lane centerlines and traffic elements.
For this work, we report results on the primary subset (\texttt{subset\_A}), which is labeled on top of the Argoverse dataset \cite{Argoverse2}. 
We adopt the metrics from the dataset's lane centerline evaluation and evaluate performance within 50m in-front and behind, and 25m to either side of the ego-vehicle. Reported metrics include DET\_l, TOP\_ll, DET\_t, and TOP\_lt \cite{wang2023openlanev2}, which correspond to the mean average precision (mAP) on directed lane centerlines, traffic elements, topology among lane centerlines, and topology between lane centerlines and traffic elements. The mAP is averaged over match thresholds of
\{1, 2, 3\}m on Frechet distance for lane centerlines and a match threshold of 0.75 IoU for traffic elements when determining correspondence between
detections and ground truths. We additionally report the 
consolidated OpenLane-V2 Score (OLS), which was released with the dataset's CVPR 2023 challenge.

\mypara{Baselines.} 
We experiment with two high-performing, open-source lane-topology models: BEVFormer-DeTR baseline and TopoNet \cite{li2023toponet}. The BEVFormer-DeTR baseline was released with the OpenLane-V2 dataset as ``baseline large", and is denoted as ``Baseline" in our experiments. It leverages the architecture from \cite{li2022bevformer} to encode multi-camera features into birds-eye-view (BEV), and decode them into lane centerlines and traffic elements using a customized Deformable-DeTR \cite{deformDetr} head for lane-topology reasoning. TopoNet is the current state-of-the-art open-sourced work, which also constructs a BEV representation of the scene, and additionally leverages a graph neural network \cite{kipf2016semisupervised} to explicitly reason about lane centerline and traffic element topology. 

\mypara{Implementation Details.} For reproducibility, we use the official implementation of both the Baseline model \cite{wang2023openlanev2} and the TopoNet model \cite{li2023toponet}. Both models use the default ResNet50 image backbone, and the output representation of the Baseline model is changed to the 11-point representation used in TopoNet for a fair comparison. Our \ours{} Transformer encoder is implemented with the official Pytorch implementation. 
All road types from the queried SD map are grouped into the following categories: pedestrian, highway, residential, service, bus\_way, and truck\_road, with an additional catch-all category. We set $N=11$ for the polyline sequence representation of SD maps, $d=32$ for the dimension of positional embeddings, and $L=6$ layers for the SD map Transformer encoder. 

\subsection{Lane Topology Prediction Performance}
We show our performance on the OpenLane-V2 dataset in \autoref{tab:olv-main}, comparing results of lane-topology models with and without \ours{}. Observe that across all metrics, we obtain a significant performance boost by using SD map information as compared to models without. For both Baseline and TopoNet, adding \ours{} gives about 15\% performance boost to the consolidated OLS metric. 
In particular, lane detection and lane-topology prediction, corresponding to metrics DET\_l and TOP\_ll respectively, receive the largest boost in performance. The Baseline model and TopoNet gain 9.8 mAP and 5.2 mAP, respectively, for lane detection and both models nearly double their lane centerline-topology prediction performances. This aligns with our intuition that SD maps contain information of the road topology which is helpful to compliment onboard camera inputs to further reason lane-level topology. 

\begin{figure*}
    \centering
    \includegraphics[width=\linewidth]{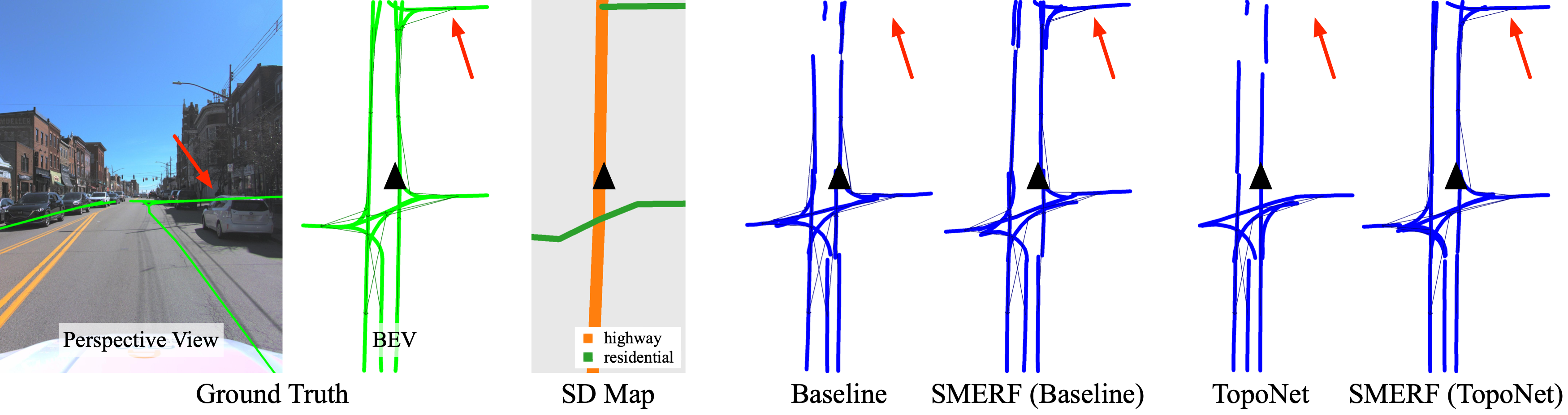}
    \caption{\textbf{Qualitative Results.} We visualize the lane predictions from OpenLane-V2 dataset validation split along with the ground truth lane-topology and the corresponding SD map from the location. Observe that adding SD maps allows the model to infer lanes that are far away and even occluded by a building (marked with the {\color{red}red} arrows). We denote the ego-vehicle's position with a black triangles for visualization purposes. 
    }
    \label{fig:qualitative}
    \vspace{-\baselineskip}
\end{figure*}

\begin{table}
\centering
\caption{\textbf{Comparison of SD map models.} In short, adding SD maps boosts lane-topology performance regardless of the methods, and the proposed \ours{} is most effective and can provide the highest performance gains. 
}
\label{tab:map-method}
\begin{tabular}{@{}llccccc@{}}
\toprule
Method &  & DET\_l & TOP\_ll & DET\_t & TOP\_lt & OLS \\ \midrule
Baseline (-- map) &  & 17.0 & 2.3 & 48.5 & 16.2 & 30.2 \\ [-\jot]
  \multicolumn{7}{@{}c@{}}{\makebox[\linewidth]{\dashrule[black!60]}} \\[-\jot]
Raster Map &  & 20.6 & 2.0 & 49.4 & 15.5 & 30.9 \\
VectorNet &  & 19.1 & 2.3 & \textbf{50.3} & 16.2 & 31.2 \\
\ours{} (Ours) &  & \textbf{26.8} & \textbf{3.9} & 48.9 & \textbf{19.2} & \textbf{34.8} \\ \bottomrule
\end{tabular}

\vspace{-15px}
\end{table}

\mypara{Performance Breakdown.}
To better understand where the performance gains are from, we break performance down by evaluating at lane centerlines that are close-by \vs far-away and intersections \vs non-intersections in \autoref{tab:olv-metrics-split}. 
When comparing performance between close-by vs. far-away areas, we observe that the majority of improvements lies in the lane centerline predictions that are further away; by adding SD maps as a prior, far-away lanes detection and topology reasoning enjoy a significant improvement. Both models' performance without the SD map prior drops significantly when attempting to reason about lane topology that are far-away---3.4 points and 4 points respectively on the OLS metric. However, by incorporating SD maps via \ours{}, the models have a much smaller performance degradation in comparison---0.8 points and 1.5 points. Surprisingly, when comparing performance at intersections \vs non-intersections, we observe that improvements are more significant for lane-topology reasoning (TOP\_ll metric). This suggests that SD maps contain priors that are useful to reason about more complicated scenes such as intersections. 

\mypara{Qualitative Comparison.}
In \autoref{fig:qualitative} from left to right, we visualize the frontal camera view, ground truth lane-topology, the corresponding SD map of the area, and lane-topology predictions of the baseline models with and without \ours{}. Observe that using the SD map allows the models to better reason and predict lane centerlines further away, especially in the case where the information is missing in the camera images due to occlusion or far range. This observation aligns with our quantitative results and provides an intuition for why SD maps help most in these challenging cases, which could be crucial to long-horizon behavior planning. 

\subsection{Analysis and Ablation Study}

We analyze the information provided by SD maps and ablate components of \ours. All experiments are based on the Baseline provided by the official OpenLane-V2 codebase.

\mypara{Comparison of SD Map Models.}
In this work, we leverage information provided in SD maps to improve lane-topology prediction. 
An astute reader may wonder if SD map information can improve performance regardless of SD map representation encoders and how effective our proposed \ours{} is.
In \autoref{tab:map-method}, we compare \ours{} against two representative methods: 1) Raster Map \cite{yang2020hdnet} and 2) VectorNet \cite{2020vectornet}. Observe that using SD map boosts lane-topology performance, \textit{regardless of the method used}. Raster Maps represent the SD map as a heatmap with a one-hot representation to encode the road type; the map features are obtained using a ResNet50 backbone. While it is able to improve lane detection, such a representation struggles with relational reasoning and performs poorly on metrics of TOP\_ll and TOP\_lt. VectorNet represents the SD map as a graph of polylines and uses a Graph Neural Network to refine the map features. Thus, it is better at relational reasoning. Overall, \ours{} shows the strongest performance since its architecture is designed explicitly for SD map encoding.

{\setlength{\tabcolsep}{4px}
\begin{table}
\centering
\caption{\textbf{Ablation on the road-type information.} Observe that the addition of semantic information (road types) from SD maps is helpful to improve the performance of \ours{}. 
}
\label{tab:lanetype-inclusion}
\begin{tabular}{@{}ccccccccc@{}}
\toprule
\multicolumn{3}{c}{Road-Types} &  & \multirow{2}{*}{DET\_l} & \multirow{2}{*}{TOP\_ll} & \multirow{2}{*}{DET\_t} & \multirow{2}{*}{TOP\_lt} & \multirow{2}{*}{OLS} \\ \cmidrule(r){1-3}
Road & Ped-Way & Others &  &  &  &  &  &  \\ \midrule
-- & -- & -- & & 17.0 & 2.3 & 48.5 & 16.2 & 30.2 \\ [-\jot]
  \multicolumn{9}{@{}c@{}}{\makebox[\linewidth]{\dashrule[black!60]}} \\[-\jot]
$\checkmark$ &  &  &  & 20.6 & 3.7 & 51.5 & 18.0 & 33.4 \\
$\checkmark$ & $\checkmark$ &  &  & 22.7 & 3.8 & 52.3 & 19.8 & 34.8 \\
$\checkmark$ & $\checkmark$ & $\checkmark$ &  & 26.8 & 3.9 & 48.9 & 19.2 & 34.8 \\ \bottomrule
\end{tabular}
\vspace{-0.3cm}
\end{table}}

\mypara{Analysis on Road Types.} 
We analyze how the road type information contributes to the performance. 
We first group the types of roads into three high-level categories: (1) roads (\ie highway and residential), (2) pedestrian-way (pedestrian walkways, crosswalks, \etc), and (3) other types (bus\_way, truck\_road, service, \etc). Then, we report performance in \autoref{tab:lanetype-inclusion} by incrementally adding the road type information. Observing that adding more priors of semantic information (\ie road types) improves performance. Even with only labels for ``roads'' present in SD maps, the Baseline model is able to achieve a significant performance gain.

\begin{table}[]
\centering
\caption{\textbf{Results on geo-disjoint training and validation split.} We re-split the training and validation set to be geographically disjoint. Observe that adding \ours{} still provides a consistent performance boost, despite the models being evaluated on a more challenging data split.}
\label{tab:disjoint-main}
\begin{tabular}{@{}llccccc@{}}
\toprule
 &  & DET\_l & TOP\_ll & DET\_t & TOP\_lt & OLS \\ \midrule
Baseline &  & 5.4 & 0.3 & 41.0 & 5.7 & 16.9 \\
Baseline+\ours{} &  & 8.8 & 0.5 & 46.3 & 6.9 & 22.1 \\ [-\jot]
  \multicolumn{7}{@{}c@{}}{\makebox[\linewidth]{\dashrule[black!40]}} \\[-\jot]
$\triangle$ Improvement &  & {\color[HTML]{34A853} 64.6\%} & {\color[HTML]{34A853} 66.7\%} & {\color[HTML]{34A853} 13.0\%} & {\color[HTML]{34A853} 21.4\%} & {\color[HTML]{34A853} 31.1\%} \\ \midrule
TopoNet &  & 14.9 & 1.0 & 34.3 & 7.6 & 21.7 \\
TopoNet+\ours{} &  & 17.0 & 1.4 & 35.4 & 8.6 & 23.4 \\ [-\jot]
  \multicolumn{7}{@{}c@{}}{\makebox[\linewidth]{\dashrule[black!40]}} \\[-\jot]
$\triangle$ Improvement &  & {\color[HTML]{34A853} 14.4\%} & {\color[HTML]{34A853} 31.2\%} & {\color[HTML]{34A853} 3.1\%} & {\color[HTML]{34A853} 14.0\%} & {\color[HTML]{34A853} 7.5\%} \\ \bottomrule
\end{tabular}

\vspace{-4px}
\end{table}

\mypara{Evaluation on Geo-disjoint Data Split.} 
The standard training and validation split in the OpenLane-V2 dataset consists of geographically overlapping areas, and models trained and tested under such a setting are known to overfit \cite{2021ConfidencePseudoLidar,li2023toponet}. We analyze the performance on a \textit{geographically disjoint} training and validation split and report our results in \autoref{tab:disjoint-main}. Observe that improvements with the SD map are consistent and significant even in the geographically disjoint splits, despite the performance drops in both the Baseline and TopoNet on this more challenging data split. Future work on these geo-disjoint splits is suggested for reducing overfitting and improving generalization across geographical areas.

\mypara{Effects on Different Model Components.}
We additionally ablate components of our model to justify our implementation details. 
We report ablation results when adding different Transformer architecture components in \autoref{tab:transf-components}. Adding the transformer model naively gives only a small boost in performance. By encoding the x,y positions of the SD map polylines with sine-cosine encoding \cite{vaswani2023attention}, we gain expressivity for the model to represent the locations of the map elements, thus boosting performance. Lastly, normalizing point coordinates into a range of $(0, 2\pi)$ prior to the positional encodings gives a final boost in lane detection performance.  

We additionally ablation the \ours{} architecture design and the effect of different numbers of heads on performance in \autoref{tab:transf-heads}. Observe that the dimension, which is by default quite small, performs best with fewer heads since the dimension-per-head is reduced as the number of heads is increased. Due to dataset size, model performance on lane detection is sensitive to the per-head dimension.

{
\begin{table}[]
\centering
\caption{\textbf{Ablation on Transformer encoder components.} We report performance after adding each component of the map transformer. Adding positional encoding is crucial for good topology reasoning.}
\label{tab:transf-components}
\begin{tabular}{@{}llccccc@{}}
\toprule
 &  & DET\_l & TOP\_ll & DET\_t & TOP\_lt & OLS \\ \midrule
Baseline (-- map) &  & 17.0 & 2.3 & 48.5 & 16.2 & 30.2 \\ [-\jot]
  \multicolumn{7}{@{}c@{}}{\makebox[\linewidth]{\dashrule[black!60]}} \\[-\jot]
~~+ map transformer &  & 18.2 & 2.6 & 48.1 & 16.8 & 30.9 \\
~~+ x,y-pos encoding &  & 20.0 & \textbf{3.9} & \textbf{49.6} & 18.9 & 33.2 \\
~~+ normalization &  & \textbf{26.8} & \textbf{3.9} & 48.9 & \textbf{19.2} & \textbf{34.8} \\ \bottomrule
\end{tabular}
\vspace{-0.3cm}
\end{table}
}
\begin{table}[]
\centering
\caption{\textbf{Ablation on Transformer Components.} We note that using fewer heads actually improves performance, as the per-head dimension gets larger.}
\label{tab:transf-heads}
\begin{tabular}{@{}clccccc@{}}
\toprule
\# Heads &  & DET\_l & TOP\_ll & DET\_t & TOP\_lt & OLS \\ \midrule
4 &  & 26.8 & 3.9 & 48.9 & 19.2 & 34.8 \\
8 &  & 22.8 & 3.6 & 52.5 & 19.0 & 34.5 \\
16 &  & 21.3 & 4.1 & 50.4 & 19.3 & 33.9 \\ \bottomrule
\end{tabular}

\vspace{-10px}
\end{table}
\section{Discussion and Conclusion}
\label{sec:conclusion}
In this work, we explore the benefits of leveraging readily available and cost-effective SD maps, and study how they can improve online map prediction and lane-topology reasoning. Our method, \ours, demonstrates consistent performance improvements over various lane-topology models, indicating the versatility of this approach.
Our work is the first to approach task of lane-topology prediction by leveraging SD map priors; further advancements may come from refining the representation learning process for SD maps within the Transformer architecture, potentially enabling  more significant performance gains.


\addtolength{\textheight}{-3cm}   







\bibliographystyle{IEEEtran}
\bibliography{main}


\end{document}